\documentclass[conference]{IEEEtran}
\IEEEoverridecommandlockouts
\usepackage{cite}
\usepackage{amsmath,amssymb,amsfonts}
\usepackage{algorithmic}
\usepackage{graphicx}
\usepackage{textcomp}
\usepackage{xcolor}
\def\BibTeX{{\rm B\kern-.05em{\sc i\kern-.025em b}\kern-.08em
    T\kern-.1667em\lower.7ex\hbox{E}\kern-.125emX}}
\begin{document}

\title{Channel Attention-Guided Cross-Modal Knowledge Distillation for Referring Image Segmentation\\
}

\author{\IEEEauthorblockN{Chen Yang}
\IEEEauthorblockA{
School of Information Science and Technology, TaiShan University, Taian, China \\
}
}

\maketitle

\begin{abstract}
Referring image segmentation (RIS) requires accurate segmentation of target regions in images according to language descriptions, which is a cross-modal task integrating vision and language. Existing RIS methods typically employ large-scale vision and language encoding models to improve performance, but their enormous parameter size severely restricts deployment in scenarios with limited computing resources. To solve this problem, this paper proposes a channel attention-guided cross-modal knowledge distillation method, which transfers the high-order fine-grained correlations between vision and language learned by the teacher network, as well as the correlations between semantic components represented by each channel, to the student network. Compared with the traditional pixel-wise relational distillation, this method not only enables the student to learn the knowledge of the teacher, but also retains part of its independent learning ability, alleviating the transfer of learning bias. Experimental results on two public datasets show that the proposed distillation method does not introduce additional parameters during inference and can achieve significant performance improvement for the student model.
\end{abstract}


\section{Introduction}
In recent years, with the rapid development of the field of multimodal learning, the image segmentation task has no longer been confined to the traditional paradigm that relies solely on visual input, but has gradually expanded to a complex perspective integrating multi-source information such as language and vision. Against this background, referring image segmentation has emerged, which requires models to accurately locate and segment corresponding visual regions in images based on natural language descriptions, greatly advancing cross-modal understanding between vision and language.
 
To address this task, early methods~\cite{hu2016segmentation} employed CNNs and LSTMs to extract visual and linguistic features respectively, and fused them directly through concatenation and convolutional layers. Evidently, such brute-force fusion fails to capture fine-grained correlations between vision and language, leading to inaccurate target localization in scenarios demanding high-order reasoning. To overcome these limitations, subsequent studies explored more effective vision-language interaction mechanisms. To date, a series of approaches based on attention mechanisms~\cite{shi2018key}, dynamic filtering~\cite{margffoy2018dynamic}, multi-task learning~\cite{cheng2025weakmcn}, etc., have been developed to enhance cross-modal information selection and alignment during feature fusion, thereby improving the accuracy and robustness of referring image segmentation models. However, all these methods are designed under the fully supervised paradigm, which entails an extremely heavy workload for pixel-level annotation of training samples. To alleviate this bottleneck, some recent studies have started to explore weakly supervised~\cite{liu2023referring} (e.g., using only bounding boxes or text descriptions) or zero-shot~\cite{wang2025iterprime} for referring image segmentation, with the goal of reducing reliance on dense manual annotations. In addition, some works~\cite{lai2024lisa,shang2024prompt} leverage large-scale pre-trained models such as the Segment Anything Model (SAM) and Large Language Models (LLMs), utilizing their strong generalization and semantic understanding capabilities to further boost cross-modal segmentation performance under limited or even no annotation conditions. These new ideas have injected fresh vitality into the advancement of referring image segmentation.

Nevertheless, to maximize segmentation performance, the above methods all adopt large-scale visual and language encoders (such as Swin-base, BERT-base, etc.). Although such large models deliver excellent performance in accuracy, their enormous parameter size and computational overhead impose stringent requirements on resource consumption during inference and practical deployment devices (e.g., edge computing platforms, mobile terminals, etc.), severely restricting their application in resource-constrained scenarios.
To tackle this issue, this paper proposes a cross-modal knowledge distillation method for referring image segmentation, which introduces knowledge distillation into the multimodal task of referring image segmentation. The proposed method transfers the fine-grained correlations between visual and linguistic features learned by the teacher network to the student network. The student network employs lightweight visual and language encoders, and gradually learns to accurately locate target regions under the guidance of cross-modal alignment knowledge from the teacher network, while maintaining low computational overhead. Experimental results on multiple public datasets demonstrate that the proposed method can significantly reduce the number of model parameters while achieving segmentation accuracy comparable to fully supervised large models, providing a feasible solution for deploying referring image segmentation in resource-constrained environments.

\section{Related Work}

Hu et al.~\cite{hu2016segmentation} first formally defined the task and proposed an end-to-end CNN-LSTM framework, providing an initial solution for cross-modal image segmentation. However, limited by the model design at the time, the framework used simple feature concatenation and convolutional layers for fusion, making it difficult to capture fine-grained correspondences between vision and language, especially in scenarios requiring complex spatial relationships or attribute reasoning, where segmentation accuracy was significantly insufficient. Shi et al.~\cite{shi2018key}, considering the varying contributions of different words in a sentence to the final task, proposed a key-word attention network that adaptively selects visual description words highly relevant to the target region, enhances their feature representation, and suppresses irrelevant or noisy words. Ye et al.~\cite{ye2019cross} used self-attention to model the correlations of multimodal features formed by mixing each word with visual features, achieving finer alignment between semantic fragments in the language description and specific regions in the image. Wang et al.~\cite{wang2022cris} introduced the CLIP model to referring image segmentation for the first time, designing a text-pixel contrastive learning framework that generates segmentation masks aligned with text embeddings, effectively improving cross-modal segmentation performance. Yang et al.~\cite{yang2022lavt} proposed a Language-Aware Vision Transformer that performs early fusion of language and visual features in the intermediate layers of the visual Transformer encoder, leveraging the powerful feature extraction ability of the visual encoder to achieve leading results on multiple datasets. Liu et al.~\cite{liu2023caris} pointed out from a context-aware perspective that purely aligning language and visual pixels ignores rich contextual information in images. To this end, they proposed a framework based on bidirectional cross-modal attention and introduced two sets of learnable prompts to mine alignment information between visual context and non-target pixels, achieving the best performance on multiple public benchmarks at that time. All the above schemes mainly focus on modeling fine-grained high-order correlations between vision and language for multi-modal reasoning, and operate mainly in the fully supervised paradigm.
With the emergence of large-scale pre-trained models such as CLIP and SAM, large-model-based RIS methods have gradually become a research hotspot. Meanwhile, multi-modal large language models (MLLMs) combine vision models with language models and demonstrate strong capabilities in complex vision-language reasoning tasks. Lai et al.~\cite{lai2024lisa} introduced a special [SEG] token to effectively combine an MLLM with a segmentation model, endowing the large model with referring segmentation capability. Shang et al. [8] proposed Prompt-RIS, which connects CLIP and SAM end-to-end and introduces cross-modal prompt learning and instance contrastive learning, achieving fine-grained text-pixel alignment and significant performance gains on complex RIS .
To save data annotation costs, some methods adopt weakly supervised paradigms. For example, Liu et al.~\cite{liu2023referring} proposed a text-guided weakly supervised RIS framework that optimizes the image-text response map by distinguishing positive and negative text expressions, and generates high-quality pseudo-labels using bilateral prompts and a background correction strategy, effectively improving segmentation performance. Feng et al. used object bounding boxes as annotation, learning rough object contours via a relaxed adversarial boundary loss and constructing pseudo ground truth masks. Cheng et al.~\cite{cheng2025weakmcn}, from a multi-task collaboration perspective, unified weakly supervised referring expression comprehension and segmentation. They used the comprehension branch as a teacher to guide the segmentation branch and designed dynamic visual feature enhancement and collaborative consistency modules to promote cross-task synergy. Wang et al. [6] proposed an iterative Grad-CAM refinement strategy and a primary word emphasis module, significantly improving the model's ability to handle position-sensitive text inputs and complex semantic relationships in zero-shot referring images. Besides reducing annotation cost, some methods employ fine-tuning strategies to reduce training computation. Xu~\cite{xu2023bridging} and Huang et al.~\cite{huang2025densely} used lightweight adapter modules to bridge frozen vision and language encoders, achieving cross-modal alignment without updating the main parameters of the pre-trained models, thus significantly reducing training costs while maintaining excellent segmentation performance. However, the common drawback of all the above methods is that they adopt large-scale vision and language encoding models for input feature extraction to maximize performance, which inevitably increases deployment cost.

\begin{figure*}[t]
\vspace{0mm}
\includegraphics[width=1.0\linewidth]{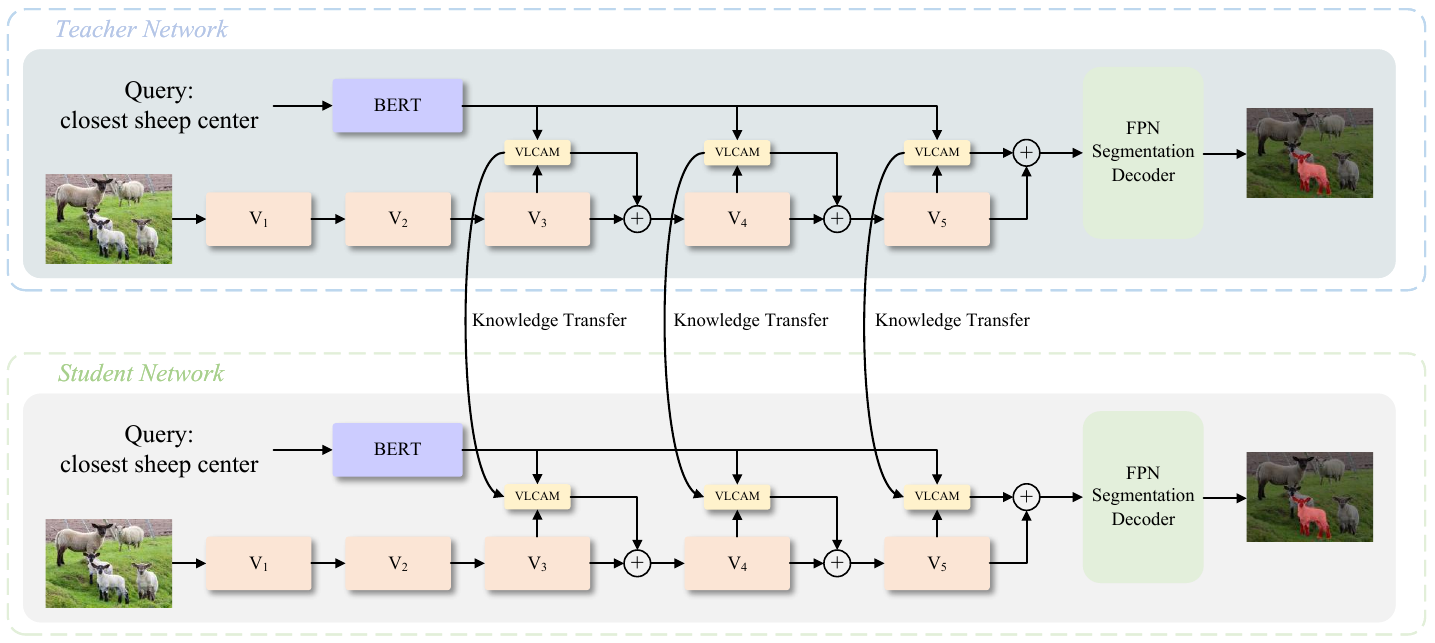}
\ \\
{\begin{center}
\vspace{-12mm}
\caption{\small{The overall architecture of the proposed model.}}
\label{fig:fig1}
\end{center}
}
\vspace{-3mm}
\end{figure*}

\begin{figure}[t]
\vspace{0mm}
\includegraphics[width=0.8\linewidth]{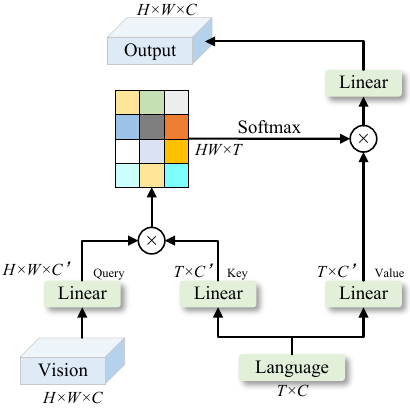}
\\
{\begin{center}
\vspace{-6mm}
\caption{\small{The vision-language cross-modal attention module.}}
\label{fig:fig2}
\end{center}
}
\vspace{-6mm}
\end{figure}

\section{Proposed Method}

The overall network architecture of the proposed method is illustrated in the Fig.~\ref{fig:fig1}, which mainly consists of a teacher network, a student network, and a cross-modal knowledge distillation mechanism. 

\subsection{Overall Network Structure}

Specifically, ResNet101 and a 12-layer BERT are used as the visual and language encoders of the teacher network. For the visual branch, the output features of each stage are denoted as $V^t\mathrm{=[}V_1^t{,}V_2^t{,}V_3^t{,}V_4^t{,}V_5^t{]}$, where $V_i^t\mathrm{\in}{R^{H_i^t{\times}W_i^t{\times}C_i^t}}$, with $H_i^t$, $W_i^t$, and $C_i^t$ representing the height, width, and number of channels of the visual features, respectively. For the language branch, the output features from the last Transformer layer are denoted as $T^t{\in}{R^{L{\times}{D^{{t}}}}}$, where $L$ is the length of the referring expression sentence and ${D^{{t}}}$ denotes the hidden dimension. The student network adopts lightweight encoders: ResNet-18 or EfficientNet-B0 for the visual branch, and the first four Transformer layers of BERT for the language branch. Their corresponding output features are denoted as $V_i^s{\in}{R^{H_i^s{\times}W_i^s{\times}{C_i^s}}}$ and $T^s{\in}{R^{L{\times}{D^{{s}}}}}$, respectively.

Next, to achieve fine-grained interaction and fusion between vision and language, this paper proposes a vision-language cross-modal attention module (as shown in the Fig.~\ref{fig:fig2}), which is applied separately to the features of the visual encoders ${V_3}\sim{V_5}$, thereby enabling progressive fusion of multimodal features. For convenience of description, let $V{\in}{R^{H{\times}W{\times}C}}$ and $T{\in}{R^{L{\times}{D}}}$ denote the visual and textual features to be fused, respectively.
Specifically, the visual feature ${V_3}$ is first reshaped into a matrix of size $HW{\times}C$, and projected to a fixed channel dimension $C_{proj}$ via a linear layer. Meanwhile, the textual feature is projected through a linear layer to form the `key' and `value', while the visual feature serves as the `query'. The calculation formulas are as follows:
\begin{equation}
V_Q=\mathrm{Linear}(V),
\end{equation}
\begin{equation}
T_K=\mathrm{Linear}(T),
\end{equation}
\begin{equation}
T_V=\mathrm{Linear}(T),
\end{equation}
where Linear denotes the linear projection layer that maps the input features to the same dimension $C'$. Subsequently, the correlation matrix $A\in R^{HW\times T}$ between vision and language can be computed as:
\begin{equation}
A=V_Q \cdot (T_K)^\top.
\end{equation}
Then, matrix $A$ is used to update the `value' to obtain the final output $O\in R^{HW{\times}C'} $ :
\begin{equation}
O=A \cdot (T_V).
\end{equation}
Finally, the output feature $O$ is integrated into the visual encoder in a residual manner ($\stackrel{.}V= V + \mathrm{Linear}(O)$) to complete the final fusion. For the decoding stage, a commonly used Feature Pyramid Network (FPN) in image segmentation is adopted to progressively fuse the multi-scale features from the encoder and predict the final segmentation map.

\begin{table*}[t]
\centering
\caption{{Performance Evaluation with distillation framework.}} 
\renewcommand{\arraystretch}{1.25}
\begin{tabular}{l|c|c|c|c|c|c}
\hline
Backbone &RefCOCO-val &RefCOCO-testA &RefCOCO-testB  &RefCOCO+-val &RefCOCO+-testA &RefCOCO+-testB   \\
\hline
ResNet-101      &69.38   &71.86   &65.02   &56.59   &62.32   &48.78  \\
\hline
ResNet-18       &61.49   &54.02   &58.26   &48.23   &53.23   &40.79    \\
w/ distill      &63.36   &65.91   &61.14   &51.82   &54.76   &44.41  \\
\hline
\end{tabular}
\label{tab:performance}
\end{table*}

\subsection{Vision-Language Relational Distillation}

In Fig. 1, both the teacher and student networks share the same overall architecture design, with the main difference lying in the number of parameters of their visual and language encoders.
To enable the student network to achieve stronger feature extraction and multimodal information modeling capabilities, this paper proposes a cross-modal feature-based knowledge distillation strategy. In referring image segmentation, a key challenge is how to model the correspondence between vision and language for multimodal reasoning. Therefore, we first adopt the MSE loss to perform knowledge guidance from the vision-language correlation matrix $A^t$learned by the teacher network to the correlation matrix $A^s$ learned by the student network:
\begin{equation}
L_{VL}=\frac{1}{HW{\times}T}   \sum_{i=1}^{HW}  \sum_{j=1}^{T} \| A^t_{(i,j)} - A^s_{(i,j)}  \| _2^2 
\end{equation}
The above loss encourages the student network to mimic the correlation modeled by the teacher network. Since the correlation is computed from multimodal features, it implicitly guides the student network to project multimodal features into the same feature space as the teacher network.

\subsection{Channel Attention Relational Distillation}

Previous studies have shown that in high‑level semantic feature maps, each channel corresponds to a certain semantic component in the image. Accordingly, this paper expects the student network to learn semantic components with strong discriminability and separability, just as the teacher network does. To this end, we introduce we introduce a soft constraint mechanism based on channel attention relational distillation, instead of directly applying pixel-level constraints, as the latter would overly restrict the learning process of the student network and hinder its ability to autonomously learn effective feature representations. This module guides the student network to mimic the teacher network's semantic discrimination capability, ensuring that it captures meaningful and distinguishable semantic components without being constrained by low-level pixel details, thereby achieving more flexible and efficient knowledge transfer. 

Specifically, this module is applied to the multi-scale features $D^t$ and $D^s$ output by the FPN-based decoder networks of the teacher and student segmentation models. Both features are first reshaped into $HW{\times}C$ followed by computing the inter-channel correlation matrix of the teacher feature as:
\begin{equation}
A_c^t= {D_t}^\top \cdot  {D_t}, \ \ A_c^s= {D_s}^\top \cdot  {D_s}
\end{equation}
Finally, we employ the mean squared error (MSE) loss to accomplish the knowledge transfer by minimizing the distance between $A_c^t$ and $A_c^s$:
\begin{equation}
L_C=\frac{1}{C{\times}C}   \sum_{i=1}^{C}  \sum_{j=1}^{C} \| A^t_{c,(i,j)} - A^s_{c,(i,j)}  \| _2^2.
\end{equation}
The total loss of the distillation network can be summarized as:
\begin{equation}
L_{d}=L_{seg} + \lambda_1 \cdot L_{VL} +  \lambda_2 \cdot L_C
\end{equation}
where $L_{seg}$ denotes the commonly used cross-entropy loss in segmentation networks, $\lambda_1$ and $\lambda_2$ are balancing hyperparameters.

\section{Experiments}

\subsection{Experimental Setup}
We evaluate our method on two public referring image segmentation benchmarks: RefCOCO~\cite{yu2016modeling} and RefCOCO+~\cite{yu2016modeling}. RefCOCO consists of 142,210 referring expressions across 19,994 images, which are split into training, validation, and test sets. RefCOCO+ follows a similar setting but prohibits the use of location-related terms such as “top” and “bottom”, focusing more on object appearance descriptions and thus posing a greater challenge.
Besides, we adopt mean Intersection over Union (mIoU), the standard metric for referring image segmentation. All input images are resized to a resolution of 384$\times$384, and the maximum text length is set to 20. We use the AdamW optimizer with an initial learning rate of 2e-5, which is decayed to one-tenth after 20 epochs.The batch size is set to 16, and the hyperparameters are set as $\lambda_1$ = 0.5 and $\lambda_2$ = 0.5.All experiments are conducted on a single NVIDIA RTX 3090 GPU.

\subsection{Experimental Results}

The proposed teacher network adopts ResNet-101 and 12-layer BERT as the visual and language encoders, respectively, which is used to guide the learning of the student network (ResNet-18). The experimental results are summarized in Table~\ref{tab:performance}. It can be observed that the lightweight student network without distillation performs significantly worse than the fully supervised large-scale model. The proposed distillation method can improve the performance of the lightweight small model by 3\%–5\% in terms of mIoU, without introducing any extra computational overhead during inference.

\section{Conclusion}

This paper proposes a cross-modal knowledge distillation method for referring image segmentation. Through correlation matrix distillation and inter-channel correlation learning, the cross-modal alignment capability of the large-scale teacher network is transferred to the lightweight student network. Experimental results demonstrate that the proposed method significantly reduces the model parameters and computational overhead while maintaining segmentation accuracy comparable to that of the fully supervised large-scale model, providing a feasible solution for the practical application of referring image segmentation on edge devices.

{
\bibliographystyle{IEEEtran}
\bibliography{egbib}
}

\end{document}